%% file: main.tex
\documentclass[nonacm]{acmart}
\usepackage{graphicx}
\usepackage{tikz}
\usepackage{microtype}
\usepackage{booktabs}
\usepackage{siunitx}
\usepackage{float}
\usepackage{multirow}
\usepackage{arydshln}
\usepackage{subcaption}
\usepackage{enumitem}
\usepackage{soul}
\usepackage{textcomp}

\newcommand{\circnum}[1]{\raisebox{.5pt}{\textcircled{\raisebox{-.9pt}{\small #1}}}}

\usepackage{pifont}
\newcommand{\cmark}{\ding{51}} 

\begin{document}

\title{MultiLexNorm++: A Unified Benchmark and a Generative Model for Lexical Normalization for Asian Languages}

\author{Weerayut Buaphet}
\email{weerayut.b_s20@vistec.ac.th}
\affiliation{%
  \institution{School of Information Science and Technology (VISTEC)}
  \city{Rayong}
  \country{Thailand.}
}

\author{Thanh-Nhi Nguyen}
\email{21521232@gm.uit.edu.vn}
\affiliation{%
  \institution{University of Information Technology, Ho Chi Minh City}
  \country{Vietnam}
}
\affiliation{%
  \institution{Vietnam National University, Ho Chi Minh City}
  \country{Vietnam}
}

\author{Risa Kondo}
\email{kondo@ai.cs.ehime-u.ac.jp}
\affiliation{%
  \institution{Graduate School of Science and Engineering, Ehime University}
  \city{Ehime}
  \country{Japan}
}

\author{Tomoyuki Kajiwara}
\email{kajiwara@cs.ehime-u.ac.jp}
\affiliation{%
  \institution{Graduate School of Science and Engineering, Ehime University}
  \city{Ehime}
  \country{Japan}
}
\affiliation{%
  \institution{D3 Center, The University of Osaka}
  \city{Osaka}
  \country{Japan}
}

\author{Yumin Kim}
\email{kimym7801@cau.ac.kr}
\affiliation{%
  \institution{Chung-Ang University}
  \city{Seoul}
  \country{Korea}
}

\author{Jimin Lee}
\email{ljm1690@naver.com}
\affiliation{%
  \institution{Chung-Ang University}
  \city{Seoul}
  \country{Korea}
}

\author{Hwanhee Lee}
\email{hwanheelee@cau.ac.kr}
\affiliation{%
  \institution{Chung-Ang University}
  \city{Seoul}
  \country{Korea}
}

\author{Holy Lovenia}
\email{holy.lovenia@gmail.com}
\affiliation{%
  \institution{SEACrowd}
  \country{Indonesia}
}
\affiliation{%
  \institution{IndoNLP}
  \country{Indonesia}
}

\author{Peerat Limkonchotiwat}
\email{peerat@aisingapore.org}
\affiliation{%
  \institution{AI Singapore}
  \city{Singapore}
  \country{Singapore}
}

\author{Sarana Nutanong}
\email{snutanon@vistec.ac.th}
\affiliation{%
  \institution{School of Information Science and Technology (VISTEC)}
  \city{Rayong}
  \country{Thailand.}
}

\author{Rob Van der Goot}
\email{robv@itu.dk}
\affiliation{%
  \institution{IT University of Copenhagen}
  \city{Copenhagen}
  \country{Denmark.}
}

\renewcommand{\shortauthors}{Buaphet et al.}
\acmArticleType{Research}
\keywords{
Lexical normalization,
Multilingual and Asian languages,
Generative models,
Benchmark datasets
}

\begin{abstract}
Social media data has been of interest to Natural Language Processing (NLP) practitioners for over a decade, because of its richness in information, but also challenges for automatic processing. Since language use is more informal, spontaneous, and adheres to many different sociolects, the performance of NLP models often deteriorates. One solution to this problem is to transform data to a standard variant before processing it, which is also called lexical normalization. There has been a wide variety of benchmarks and models proposed for this task. The MultiLexNorm benchmark proposed to unify these efforts, but it consists almost solely of languages from the Indo-European language family in the Latin script. Hence, we propose an extension to MultiLexNorm, which covers 5 Asian languages from different language families in 4 different scripts. We show that the previous state-of-the-art model performs worse on the new languages and propose a new architecture based on Large Language Models (LLMs), which shows more robust performance. Finally, we analyze remaining errors, revealing future directions for this task.
\end{abstract}

\maketitle
\input{body/introduction}
\input{body/relwork}
\input{body/benchmark}
\input{body/models}

\input{body/results}
\input{body/analysis}

\section{Discussion and Future Work}
In this section, we give discussions of our approach and outline directions for future research that could further strengthen text normalization across diverse writing systems as follows.


\begin{table}
    \centering
    \renewcommand{\arraystretch}{1.1}
    \setlength{\tabcolsep}{6pt}

    \begin{tabular}{
        |c|
        S[table-format=3.2] S[table-format=3.2] S[table-format=3.2]|
        S[table-format=3.2] S[table-format=3.2] S[table-format=3.2]|
    }
    \hline
    \multirow{2}{*}{\textbf{Langs}} 
      & \multicolumn{3}{c|}{\textbf{ERR (ÚFAL)}} 
      & \multicolumn{3}{c|}{\textbf{F1 (ÚFAL)}} \\
    \cline{2-7}
      & \textbf{Base} & \textbf{Decom.} & \textbf{Latin}
      & \textbf{Base} & \textbf{Decom.} & \textbf{Latin} \\
    \hline
    JA  &    39.24 &    33.38   &    38.28 &    61.09 &    55.44 &    58.93 \\
    KO  &    -2.38 &    -126.98 &    4.76  &    25.43 &    12.27 &    25.93 \\
    TH  &    6.90  &    7.44    &    35.39 &    24.73 &    24.84 &    62.90 \\
    \hline
    \end{tabular}
    \caption{Comparison of ERR and F1 scores across non-Latin script languages using MFR and UFAL variants. 
\textit{Base} denotes the UFAL model with word-level representations. Following the procedure described in StrokeNet~\cite{wang-etal-2022-breaking}, 
\textit{Decom.} refers to the version decomposed to sub-character level or character level (for Thai), while \textit{Latin} represents the decomposed version mapped to the Latin alphabet.}
    \label{tab:analysis_stroke}
\end{table}

\begin{itemize}
    \item We adopt word-level normalization as it is directly applicable to downstream tasks such as Named Entity Recognition and Part-of-Speech tagging. Normalization, however, can also operate at multiple levels, from characters to full sentences~\cite{ding-etal-2020-self}. To address more complex cases, future work could explore sentence-level approaches, as well as evaluation metrics that link normalization quality directly to downstream performance, e.g., dependency parsing~\cite{zhang-etal-2013-adaptive} or machine translation~\cite{wang-ng-2013-beam}.
    
    \item Our analysis shows that the baseline UFAL model struggles with Asian scripts due to segmentation issues. Prior work, such as StrokeNet~\cite{wang-etal-2022-breaking}, demonstrates that sub-character representations (e.g., stroke sequences) can address this bottleneck and improve translation quality. 
    While our framework prioritizes generality across multiple writing systems, future work should consider incorporating language-specific techniques, such as stroke-level or sub-character-based models.
    To demonstrate the potential for improvement, we evaluated three non-Latin script languages using the baseline ÚFAL models and applied the StrokeNet~\cite{wang-etal-2022-breaking} approach, which decomposes text into sub-character or character units (e.g., in Thai) and maps them to Latin characters. As shown in Table~\ref{tab:analysis_stroke}, performance decreased when models were applied directly to decomposed units, which confirms our segmentation analysis in Section~\ref{ap:tokenization}.
    However, mapping the decomposed representations to Latin characters consistently improved results across all languages.
    These findings suggest that leveraging decomposed character representations and mapping non-Latin scripts to Latin-based forms can enhance model performance for languages with non-Latin orthographies.

\end{itemize}

\section{Conclusion}
When extending evaluation for lexical normalization to Asian languages, we found that state-of-the-art performs much worse, especially for languages in scripts with longer byte sizes. Hence, we propose using LLMs with in-context learning for this task; these show stronger performance overall, although they do rely on a pre-detection step and a lookup list generated from the training data. We also report on bottlenecks of the model: normalization detection, spelling errors, and slang are still open challenges for future work.
Additionally, we observed that open-source LLMs do not consistently outperform the fine-tuned ÚFAL baseline, and a clear gap remains between open- and closed-source models in text normalization.
We expect our findings to support future research on developing effective methods for language-specific and multilingual text normalization.

\begin{acks}
We would like to express our sincere gratitude to Thanh-Phong Le for his contribution to the annotation of Vietnamese data.
\end{acks}

\bibliographystyle{ACM-Reference-Format}
\bibliography{references}

\appendix

\input{body/data_statement}

\section{LLM Normalization results}
\label{llmonly}

\begin{table}[th]
    \small
    \centering
    \begin{tabular}{|l|S[table-format=2.2] S[table-format=2.2] S[table-format=2.2] S[table-format=2.2] S[table-format=2.2]|}
    \hline
    \textbf{Lang.} &
    \multicolumn{1}{c}{\textbf{MFR}} &
    \multicolumn{1}{c}{\textbf{LLaMA3}} &
    \multicolumn{1}{c}{\textbf{Qwen2.5}} &
    \multicolumn{1}{c}{\textbf{Gemma3}} &
    \multicolumn{1}{c|}{\textbf{GPT-4o}} 
    \\ \hline
        
        IN             & \textbf{58.94}         & 37.37	          & 38.58	            & 44.64	                           & \underline{53.43}\\
        JA             & 5.77                   & 6.76	          &\underline{16.45}    & 7.17                             & \textbf{20.48}\\
        KO             & \textbf{6.35}          & -19.31          &\underline{-11.64}   &-13.49                            & -12.17 \\
        TH             & \textbf{41.87}         & 18.14	          & 21.52	            &29.73	                           & \underline{37.93}\\
        VI             & \textbf{75.58}         & 35.23	          & 50.25	            &56.42	                           & \underline{72.23}\\  \hline
        DA             & 49.68                  & 60.55	          & 56.65	            &\underline{64.24}	               & \textbf{66.67}\\
        DE             & 31.95                  & 50.41	          & 56.91	            &\underline{59.61}	               & \textbf{64.49}\\
        EN             & \textbf{64.93}         & 54.17	          & 57.30               & 52.08	                           & \underline{58.62}\\
        ES             & 25.57                  & 29.80	          & 37.49	            &\underline{38.12}	               & \textbf{40.68}\\
        HR             & 36.52                  & 30.26	          & 30.84	            &\underline{44.20}	               & \textbf{46.73}\\
        ID-EN          & \textbf{61.17}         & 44.27	          & 50.80               & 50.11                            & \underline{56.59}\\
        IT             & 18.63                  & 37.29	          & 42.57	            &\underline{45.21}	               & \textbf{45.87}\\
        NL             & 37.58                  & 30.09	          & 33.36	            &\underline{44.12}	               & \textbf{57.07}\\
        SL             & \underline{56.71}      & 32.28	          & 34.55	            & 53.20                            & \textbf{63.67}\\
        SR             & 42.52                  & 42.70	          & 38.05	            &\underline{52.58}	               & \textbf{56.58}\\
        TR             & 15.42                  & 36.15	          & 34.85	            &\underline{48.37}	               & \textbf{54.73}\\
        TR-DE          & 21.82                  & 23.96	          & 26.94	            &\underline{36.20}	               & \textbf{51.62}\\   \hline
        \textbf{Avg.}  & 38.29                  & 32.36	          & 36.20	            &\underline{41.91}	               & \textbf{49.13}\\
        \hline
    \end{tabular}
    \caption{ERR scores across language settings. Bold indicates the highest (best) score per row; underlined denotes the second highest. LLM results are averaged over three runs each.}
    \label{tab:mainresults_llm_only}
\end{table}

In this setup, we aim to test how well the LLM can perform text normalization without a dictionary-based lookup. 
As shown in ~\ref{tab:mainresults_llm_only}, there is a large gap between proprietary and open-source models, only Gemma and GPT-4o can surpass the MFR base model for the average performance.
For new languages, MFR is a strong baseline model. It surpasses all LLM models in Indonesian, Korean, Thai, and Vietnamese, except Japan.
This result indicates that using LLM alone for new languages remains challenging, especially for non-Latin scripts (Japanese, Korean, Thai), where the results are substantially lower compared to Latin script languages.

\textbf{Common errors}: We perform a manual inspection of random subsamples from GPT-4o for all languages. We found some clear trends on which types of errors are being made:
\begin{enumerate}
    \item Normalizations that are context-dependent or have multiple potential normalizations, due to, e.g., politeness markers. 

    \item Spelling errors and phonetical replacements; these can be more challenging for non-Latin scripts, as they can both be more nuanced (there are more smaller diacritic-like differences, like tonal information), or have much more orthographic distance, as phonetical replacements do not necessarily have any overlap with the original.
    
    \item Transliterations of foreign words; we had corrected transliterations during annotations, but they result in words that are not normally included in vocabularies.
    
    \item For Korean, we additionally found that emotional nuance, hateful words, and missing cultural context lead to lower performance. 
\end{enumerate}

\section{Normalization detection results}
\label{appendix:detection_model}
\begin{table}[th]
    \small
    \centering
    \begin{tabular}{|l|S[table-format=2.2] S[table-format=2.2] S[table-format=+2.2]|}
        \hline
        \textbf{Language} & \textbf{MFR} & \textbf{Encoder}  & \textbf{UFAL}\\
        \hline
            IN       & 74.78  & 96.70   & 84.21             \\
            JA       & 49.46  & 78.49   & 71.03             \\
            KO       & 19.58  & 59.13   & 44.33             \\ 
            TH       & 68.82  & 73.80   & 29.13             \\
            VI       & 89.59  & 94.69   & 91.55             \\ \hline
            Average New languages & 60.45 & 80.56 & 64.05\\
            \hline
            DA       & 68.94  & 84.23   & 84.30             \\
            DE       & 58.46  & 86.87   & 84.50             \\
            EN       & 81.29  & 85.77   & 86.78             \\
            ES       & 42.48  & 76.04   & 75.24             \\
            HR       & 57.16  & 79.49   & 82.07             \\
            ID-EN    & 79.11  & 86.72   & 87.54             \\
            IT       & 45.21  & 75.26   & 76.24             \\
            NL       & 64.51  & 85.21   & 84.53             \\
            SL       & 76.55  & 88.89   & 88.93             \\
            SR       & 61.67  & 82.79   & 83.96             \\
            TR       & 33.74  & 84.64   & 85.97             \\
            TR-DE    & 45.68  & 84.45   & 86.71             \\ \hline
             Average Original languages & 59.57 & 83.36 & 83.90\\
        \hline
        \textbf{Average} & 62.37 & 83.49 & 78.06\\
        \hline
    \end{tabular}
    \caption{F1-score (\%) for informal text detection performance.}
    \label{tab:detection_performance}
\end{table}

\textit{To what extent can encoder-based models improve the detection performance of unseen informal words compared to dictionary-based approaches?}

\textbf{Setup}:  
We build detection models for 17 languages, drawing inspiration from the detection component of the Named Entity Recognition (NER) task~\cite{ma-etal-2022-decomposed}.  
Our implementation is based on the MaChAmp toolkit~\cite{van-der-goot-etal-2021-massive}, which supports multi-task and multilingual learning.  
Given the scale of our experiments, we use the standard hyperparameters of MaChAmp. The model architecture employed was XLM-Roberta-base, a multilingual transformer model. Training was conducted with a learning rate of 1e-4 and a batch size of 32. The training process ran for 20 epochs, and a dropout rate of 0.2 was applied to help prevent overfitting. These hyperparameters were selected to optimize model performance for all detection models.
Similar to the NER task, we report F1-scores (\%) based on labeled class performance, excluding tokens assigned to the non-normalized (O) class.

\textbf{Results}:
As shown in Table~\ref{tab:detection_performance}, encoder-based models improve detection performance over the dictionary-based MFR approach by approximately 24\%, and perform better than the UFAL model for new languages.
For the detection of which words to normalize, UFAL clearly performs much worse on non-Latin languages (Table~\ref{tab:detection_performance}). Our encoder-based detection model performs more robustly, but also shows that the detection task itself is still an important challenge of the full lexical normalization task, as it provides a performance upper bound.


\input{body/appendix}

\section{Performance Comparison Across Metrics for Text Normalization}
We extend our evaluation beyond ERR and F1 by presenting comparative results across languages for Precision and Recall in Table~\ref{tab:supplemental_results_precision_recall}, and for Accuracy in Table~\ref{tab:supplemental_results_accuracy}. 
%

\begin{table}
    \small
    \centering
    \setlength{\tabcolsep}{1.3pt} 
    \begin{tabular}{|l|
            S[table-format=3.2] S[table-format=2.2] S[table-format=2.2] S[table-format=2.2] S[table-format=2.2] S[table-format=2.2]|
            S[table-format=2.2] S[table-format=2.2] S[table-format=2.2] S[table-format=2.2] S[table-format=2.2] S[table-format=2.2]|
        }
        \hline
        \multirow{2}{*}{\textbf{Lang.}} & \multicolumn{6}{c|}{\textbf{Precision}} & \multicolumn{6}{c|}{\textbf{Recall}} \\
        \cline{2-13}
        & \textbf{MFR} & \textbf{ÚFAL} & \textbf{LLaMA} & \textbf{Qwen} & \textbf{Gemma} & \textbf{GPT-4o} 
        & \textbf{MFR} & \textbf{ÚFAL} & \textbf{LLaMA3} & \textbf{Qwen} & \textbf{Gemma} & \textbf{GPT-4o} 
        \\
        \hline
        IN                & 100.00    & 96.82    & 95.35    & 95.58    & 95.55     & 95.39   & 58.94	& 70.24	    & 66.48    & 67.86     & 69.72	    & 71.82 \\
        JA                & 56.29     & 84.94    & 72.50    & 77.96    & 72.22     & 77.17   & 25.84	& 47.70	    & 26.21    & 28.36     & 23.78	    & 31.98 \\
        KO                & 78.57     & 46.81    & 12.02    & 15.72    & 20.20     & 19.44   & 8.73	    & 17.46	    & 2.71	   & 3.67	   & 5.80	    & 4.83 \\
        TH                & 78.65     & 64.56    & 74.44    & 75.61    & 75.71     & 75.68   & 57.47	& 15.30	    & 61.51    & 62.35     & 63.53	    & 64.77 \\
        VI                & 96.82     & 96.52    & 95.20    & 95.51    & 95.33     & 96.17   & 78.15	& 80.24	    & 80.46    & 81.62     & 81.55	    & 83.36 \\  \hline
        DA                & 93.85     & 90.87    & 88.93    & 87.52    & 89.20     & 88.37   & 53.16	& 72.47	    & 73.73    & 71.73     & 74.68	    & 76.16 \\
        DE                & 90.83     & 89.47    & 89.28    & 89.87    & 90.87     & 91.00   & 35.7	    & 69.82	    & 54.69    & 57.62     & 59.01	    & 61.07 \\
        EN                & 94.15     & 90.22    & 90.66    & 91.01    & 90.71     & 90.62   & 69.23	& 78.96	    & 76.43    & 77.42     & 77.11	    & 77.44 \\
        ES                & 99.20     & 83.95    & 81.98    & 83.88    & 84.31     & 83.81   & 25.78	& 52.18	    & 51.70    & 56.62     & 58.63	    & 60.64 \\
        HR                & 93.40     & 92.28    & 84.25    & 85.05    & 86.81     & 86.92   & 39.30    & 66.95	    & 55.48    & 54.79     & 61.30	    & 64.88 \\
        ID-EN             & 90.27     & 83.36    & 83.94    & 84.42    & 83.75     & 83.47   & 68.56	& 81.79	    & 79.95    & 80.07     & 80.07	    & 80.41 \\
        IT                & 70.73     & 74.74    & 80.05    & 80.27    & 82.07     & 80.60   & 28.71	& 70.30	    & 53.14    & 59.08     & 58.42	    & 61.72 \\
        NL                & 95.74     & 94.70    & 90.20    & 90.37    & 91.34     & 92.27   & 39.46	& 60.34	    & 47.91    & 49.68     & 56.28	    & 61.53 \\
        SL                & 95.05     & 92.57    & 91.38    & 91.59    & 92.66     & 93.11   & 59.83	& 78.26	    & 66.58    & 66.62     & 72.44	    & 76.56 \\
        SR                & 96.4      & 93.05    & 90.52    & 90.34    & 91.90     & 92.22   & 44.27	& 69.69	    & 60.40    & 57.87     & 65.77	    & 67.99 \\
        TR                & 87.72     & 89.38    & 86.22    & 86.73    & 90.93     & 89.29   & 16.89	& 65.37	    & 47.92    & 44.88     & 56.08	    & 62.39 \\
        TR-DE             & 84.81     & 88.52    & 82.01    & 82.24    & 83.37     & 86.26   & 26.91	& 66.87	    & 45.66    & 47.77     & 52.67	    & 63.48 \\ \hline
        \textbf{Avg.}     & 88.38     & 85.46    & 81.70    & 82.57    & 83.35     & 83.63   & 43.35	& 62.58	    & 55.94    & 56.94     & 59.81	    & 63.00 \\ \hline
    \end{tabular}
    \caption{Precision and Recall (\%) across languages for the lexicon normalization task. Results for LLMs are averaged over three runs.}
    \label{tab:supplemental_results_precision_recall}
\end{table}

\begin{table}[h]
    \small
    \centering
    \begin{tabular}{|l|S[table-format=2.2] S[table-format=2.2] S[table-format=2.2] S[table-format=2.2] S[table-format=2.2] S[table-format=2.2]|}
        \hline
        \multirow{2}{*}{\textbf{Lang.}} & \multicolumn{6}{c|}{\textbf{Accuracy}}\\
        \cline{2-7}
        & \textbf{MFR} & \textbf{ÚFAL} & \textbf{LLaMA} & \textbf{Qwen} & \textbf{Gemma} & \textbf{GPT-4o}
        \\ \hline
        IN            &  80.59	& 84.85	    & 82.62	    & 83.33	    & 84.14	    & 85.04\\
        JA            &  93.93	& 96.09	    & 94.61	    & 94.87	    & 94.50	    & 95.01\\
        KO            &  92.47	& 91.77	    & 77.69	    & 77.91	    & 78.08	    & 78.06\\
        TH            &  97.61	& 96.18	    & 97.55	    & 97.63	    & 97.66	    & 97.70\\
        VI            &  96.04	& 96.33	    & 96.18	    & 96.40	    & 96.36	    & 96.77\\  \hline
        DA            &  95.77	& 97.07	    & 97.02	    & 96.76	    & 97.13	    & 97.15\\
        DE            &  88.13	& 93.29	    & 90.94	    & 91.46	    & 91.87	    & 92.14\\
        EN            &  97.23	& 97.66	    & 97.52	    & 97.61	    & 97.57	    & 97.58\\
        ES            &  94.60	& 95.81	    & 95.67	    & 96.07	    & 96.18	    & 96.30\\
        HR            &  91.84	& 95.03	    & 92.94	    & 92.95	    & 93.81	    & 94.23\\
        ID-EN         &  94.82	& 95.40	    & 95.29	    & 95.37	    & 95.27	    & 95.27\\
        IT            &  95.79	& 97.29	    & 96.96	    & 97.19	    & 97.21	    & 97.31\\
        NL            &  82.97	& 88.23	    & 84.33	    & 84.80	    & 86.48	    & 88.07\\
        SL            &  93.62	& 95.87	    & 94.15	    & 94.18	    & 95.07	    & 95.71\\
        SR            &  94.53	& 96.61	    & 95.62	    & 95.39	    & 96.14	    & 96.40\\
        TR            &  69.13	& 84.69	    & 78.42	    & 77.61	    & 82.25	    & 83.71\\
        TR-DE         &  76.20	& 87.23	    & 80.34	    & 80.89	    & 82.31	    & 85.76\\   \hline
        \textbf{Avg.} &  90.31	& 93.49	    & 91.05	    & 91.20	    & 91.88	    & 92.48\\
        \hline
    \end{tabular}
    \caption{Accuracy scores across (\%) language settings for the lexicon normalization task. LLM results are averaged over three runs each.}
    \label{tab:supplemental_results_accuracy}
\end{table}

\end{document}

%% file: body/introduction.tex
\section{Introduction}
%
It has been shown that variance in language varieties often has a negative effect on the performance of NLP models~\citep[]{gururangan-etal-2020-dont,ziems-etal-2023-multi}. In particular, the social media domain contains challenging language due to its informal, diverse, ever-changing nature~\cite{eisenstein-2013-bad,plank2016non}. There has been a variety of methods proposed for improving transfer performance to this domain, including language model adaptation~\cite{nguyen-etal-2020-bertweet,barbieri-etal-2022-xlm}, self-training~\cite{foster-etal-2011-news,liu-etal-2018-parsing}, and lexical normalization~\cite{han-baldwin-2011-lexical,eisenstein-2013-bad}. Lexical normalization is the task of transforming
an utterance into its standard form, word by word,
including both one-to-many (1-n) and many-to-one
(n-1) replacements~\cite{van-der-goot-etal-2021-multilexnorm}. 

Existing literature indicates that by making input data closer to more standard training data, performance improves for a variety of NLP tasks, including intent classification~\cite{vielsted-etal-2022-increasing}, named entity recognition~\cite{schulz2016multimodular,plank-etal-2020-dan}, POS tagging~\cite{derczynski-etal-2013-twitter,zupan2019tag},
dependency parsing~\cite{zhang-etal-2013-adaptive,baldwin-li-2015-depth,van-der-goot-van-noord-2018-modeling,van-der-goot-etal-2020-norm,van-der-goot-etal-2021-multilexnorm},
dependency parsing~\cite{zhang-etal-2013-adaptive,baldwin-li-2015-depth,van-der-goot-etal-2021-multilexnorm},
 and constituency parsing~\cite{van-der-goot-van-noord-2017-parser}, 
sentiment analysis~\citetext{\citealp{Sidarenka:19,kondo-etal-2025-text}}, 
and machine translation~\cite{bhat-etal-2018-universal,rosales-nunez-etal-2021-understanding,naplava-etal-2021-understanding}. Recent work has shown that both open large language models~\cite{aliakbarzadeh2025exploring} and proprietary models are also vulnerable to input noise~\cite{popovic-etal-2024-effects, srivastava-chiang-2025-calling}. 

Despite progress in lexical normalization, most benchmarks remain biased toward Indo-European languages with Latin script, limiting our understanding of how models perform on languages with different scripts and morphologies. 
Therefore, we investigate: 1) How well do existing normalization models generalize to Asian languages in a variety of language families and scripts? 2) How can generative language models be used for lexical normalization in a wide variety of languages?

In this work, we investigate lexical normalization beyond the Indo-European language family. We adapt existing datasets to the existing MultiLexNorm benchmark and create new datasets to increase coverage. 
Our contributions are: 1) We release MultiLexNorm++: a manually annotated extension to MultiLexNorm covering 5 language families and 4 different scripts. 2) We propose an LLM-based lexical normalization method that combines traditional heuristics with modern LLMs. 3) We evaluate and analyze the current state-of-the-art model and the LLM-based models, showing that our LLM-based model is more robust on our new languages, but it is still a challenging task.

%% file: body/relwork.tex
\section{Related Work}
Earlier work on lexical normalization was scattered, using different datasets, models, and evaluation metrics. MultiLexNorm was proposed to unify standards in dataset annotation and evaluation metrics. It includes 12 languages; however, 10 of these are Indo-European languages, and all languages in MultiLexNorm use the Latin script (see Table~\ref{tab:data2}). 

While there has been work on non-Indo-European languages for normalization, it remains more fragmented, and data is often not available or annotated for different setups. This includes normalization beyond the word level\cite{7f3453dca7c749528937b7d8f2b84273}, without context~\cite{wang-etal-2013-chinese,li-yarowsky-2008-mining}, or including transliteration~\cite{tursun-cakici-2017-noisy,bhat-etal-2018-universal}, which can all be considered different tasks from the one we are targeting in this work. In our work, we aim to re-use datasets that already exist, and complement these were possible with native speakers as annotators to increase the coverage beyond the original MultiLexNorm, while maintaining their task definition.

There have been a variety of models proposed for the lexical normalization task, including feature-based methods~\cite{van-der-goot-2019-monoise}, sequence labeling~\cite{wnut-seqlab}, machine translation~\cite{wnut-sinai,wnut-seqseq}, RNNs~\cite{lourentzou2019adapting}, and most recently, prompting~\cite{maarouf-tanguy-2025-automatic}. UFAL~\cite{wnut-ufal}, a fine-tuned encoder-decoder model, has been shown to outperform other models on the MultiLexnorm benchmark with a substantial  (\textasciitilde 14 p.p.) margin ~\cite{van-der-goot-etal-2021-multilexnorm} over competitors, which is why we use it as our baseline. We refer to~\citet{aliero2023systematic} for an overview of other models.

\begin{table}
    \centering
    \small
    \begin{tabular*}{\textwidth}{@{\extracolsep{\fill}} l l l r c c S[table-format=2.2] l l}
        \toprule
        \textbf{Language} & \textbf{Code} & \textbf{Source} & \textbf{Words} & \textbf{1-n/n-1} & \textbf{Caps} & \textbf{\%Norm.} & \textbf{Script} & \textbf{Family}\\
        \midrule
        \multicolumn{8}{l}{\textit{New Languages}} \\
        Indonesian &   ID & \citet{kurnia2020statistical}               & 48,716    & -         & \cmark & 47.47 & Latn & aust1307\\ 
        Japanese &     JA & New                                         & 95,411    & \cmark    & NA      & 7.03  & Jpan & japo1237\\ 
        Korean &       KO & New                                         & 16,618    & \cmark    & NA      & 7.52 & Kore & kore1284\\ 
        Thai &         TH & \citet{limkonchotiwat-etal-2021-handling}   & 169,751   & \cmark    & NA      & 4.72  & Thai & taik1256 \\ 
        Vietnamese &   VI & \citet{nguyen-etal-2024-vilexnorm}          & 128,685   & \cmark    & -      & 15.98 & Latn & aust1305\\ 
        \midrule
        \multicolumn{8}{l}{\textit{Original Languages}} \\
        Danish &         DA & \citet{plank-etal-2020-dan}                 & 20,206    & \cmark    & \cmark & 9.25  & Latn & indo1319\\ 
        German &         DE & \citet{sidarenka2013rule}                   & 24,948    & \cmark    & \cmark & 17.96 & Latn & indo1319\\ 
        English &        EN & \citet{baldwin-etal-2015-shared}            & 73,806    & \cmark    & -      & 6.90  & Latn & indo1319\\ 
        Spanish &        ES & \citet{alegria2013introduccion}             & 13,824    & -         & -      & 7.69  & Latn & indo1319\\ 
        Croatian &       HR & \citet{11356/1170}                          & 89,052    & -         & -      & 8.89  & Latn & indo1319\\ 
        Indonesian-EN &  ID-EN & \citet{barik-etal-2019-normalization}    & 23,124    & \cmark    & -      & 12.16 & Latn & aust1307\\ 
        Italian &        IT & \citet{van-der-goot-etal-2020-norm}         & 14,641    & \cmark    & \cmark & 7.32  & Latn & indo1319\\ 
        Dutch &          NL & \citet{dutchNorm}                           & 21,657    & \cmark    & \cmark & 28.29 & Latn & indo1319\\ 
        Slovenian &      SL & \citet{11356/1123}                          & 75,276    & -         & -      & 15.62 & Latn & indo1319\\ 
        Serbian &        SR & \citet{11356/1171}                          & 91,738    & -         & -      & 7.65  & Latn & indo1319\\ 
        Turkish &        TR & \citet{colakoglu-etal-2019-normalizing}     & 8,082     & \cmark    & \cmark & 37.02 & Latn & turk1311\\ 
        Turkish-German & TR-DE & \citet{van-der-goot-etal-2021-lexical}   & 16,508    & \cmark    & \cmark & 24.14 & Latn & indo1319\\ 
        \bottomrule
    \end{tabular*} 
    \caption{Dataset statistics including original sources and percentage of token normalization. Script codes are from ISO 15924, and language family codes from Glottolog.}
    \label{tab:data2}
\end{table}

%% file: body/benchmark.tex
\section{MultiLexNorm++}
We use convenience sampling for our language selection for MultiLexNorm++: we only used languages for which existing datasets existed (which we convert) or where we had access to in-house annotators. 
Below we provide descriptions of each language and dataset, more information is available in the attached dataset statement (see Appendix~\ref{app:datastatement}).
%

\subsection{Languages}
\begin{figure}[tp]
    \centering
    \includegraphics[width=0.8\linewidth]{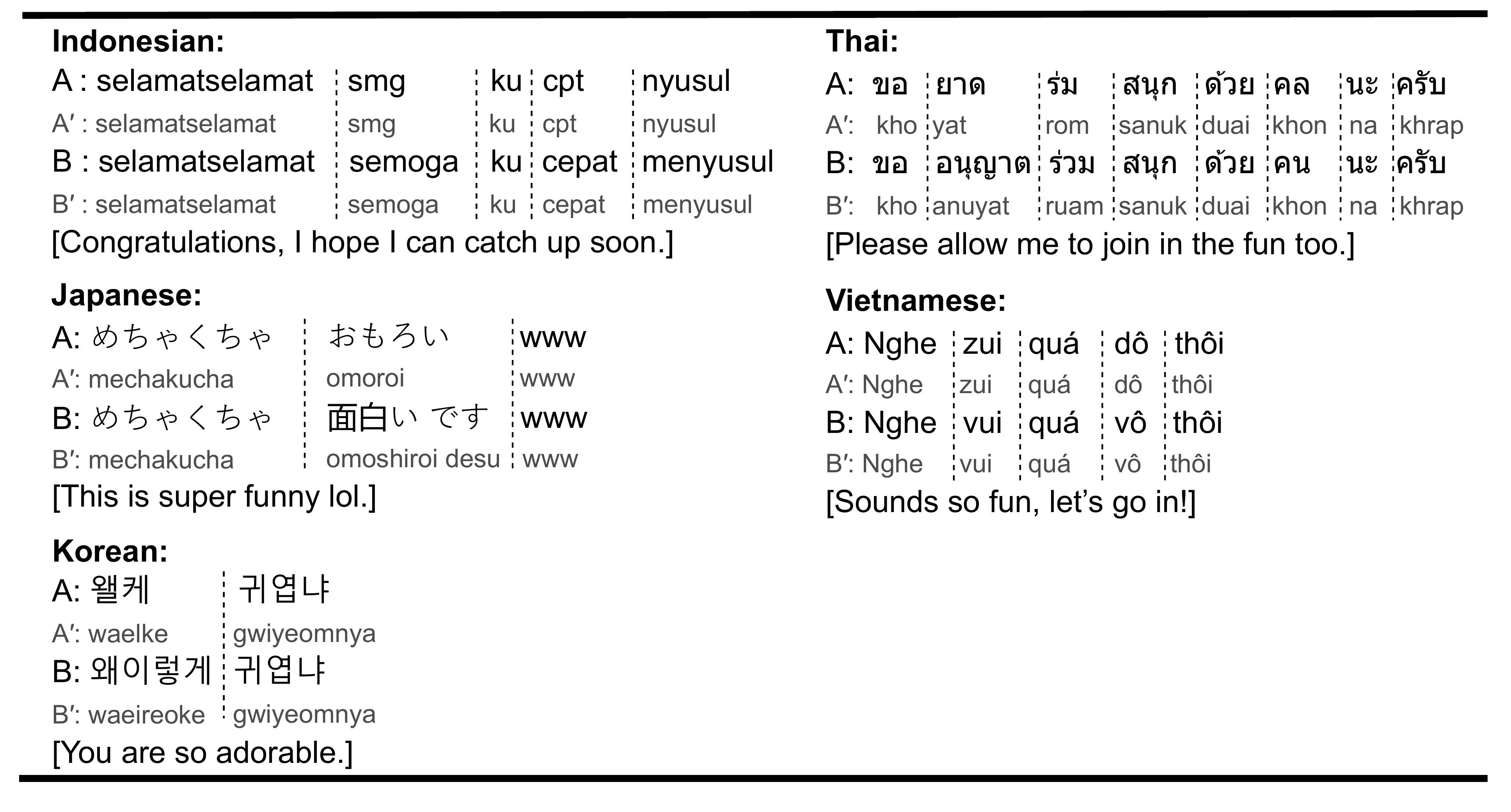}
    \caption{Normalization examples across languages. A shows the original informal or non-standard text, with A (prime) providing its romanized transliteration. B shows the normalized form, with B (prime) its romanized transliteration. English translations are provided in square brackets.}
    \label{fig:examples}
\end{figure}

As shown in Table~\ref{tab:data2}, we provide a summary of our benchmark, including new languages and original languages.
The new languages span 5 language families and 4 scripts (Figure~\ref{fig:examples}), and we describe each corresponding dataset below:
\paragraph{Indonesian}
Indonesian uses the Latin script, but in contrast to most Indo-European languages, it has a flexible word order, and tense is not marked. We use the CIL dataset~\cite{septiandri2017detecting,kurnia2020statistical}, which consists of Instagram comments and was already annotated. Following the MultiLexNorm definition of the task, we keep only the sentences that include word alignment.

\paragraph{Japanese}
Japanese is agglutinative and is commonly written using a combination of three scripts, so it has a high character diversity.  
We extracted lexical normalization instances from the WRIME~\cite{kajiwara-etal-2021-wrime, suzuki-etal-2022-japanese, kondo-etal-2025-text}, a text normalization dataset for sentiment analysis on Twitter (2010-2020).
The dataset contains about 3k tweets with 98276 (before normalization) and 100760 (after) tokens. The Japanese dataset involves both splits and merges. Normalizations involving sentence splitting or word reordering have been excluded.
Annotators are native Japanese speakers, one university teacher and two university students, the guidelines from MultiLexNorm were used.

\paragraph{Korean}
Korean is agglutinative and uses a unique writing system that represents a syllabic structure, which consists of a consonant and a vowel. 
We use the data collected by~\citet{choi2023largescale}, which was originally taken from ``dcinside.com'' (a Korean community similar to Reddit). 
Three native Korean speakers (one University professor, two students) annotated the data and obtained a kappa score of 0.916.

\paragraph{Thai}
Thai is a tonal language with a non-linear script, where tone marks and diacritics are essential to meaning. It lacks spaces between words and employs context-dependent grammar, relying on particles and temporal cues to express tense, number, or aspect.
We use 2,500 sentences randomly sampled from~\citet{limkonchotiwat-etal-2021-handling}, which consists of Twitter data (2017-2019), and has manual annotation for lexical normalization. 

\paragraph{Vietnamese} 
Vietnamese uses the Latin script augmented with an intricate system of diacritics. 
We use the normalization annotation from~\citet{nguyen-etal-2024-vilexnorm}, which was collected from social media platforms such as Facebook and TikTok. This data had an inter-annotator agreement of 88.46\%. We manually align the annotation to MultiLexNorm’s word-level format.

%% file: body/models.tex
\section{Evaluation Setup}
We evaluate a simple statistical baseline, the current state-of-the-art model, and our proposed LLM-based model on MultiLexNorm++.

\subsection{Baselines}

We use the same baseline as the MultiLexNorm paper~\cite{van-der-goot-etal-2021-multilexnorm}, and the SOTA model on their data.
\textbf{Most-Frequent-Replacement (MFR)} is a statistical-based baseline that normalizes the input word with the most frequent replacement word based on the statistics from the training set.
\textbf{ÚFAL}~\cite{samuel-straka-2021-ufal} is the state-of-the-art model for this task with at least 14 percentage points improvements over other models (on original MultiLexNorm). 
The model is a fine-tuned encoder-decoder ByT5~\cite{xue-etal-2022-byt5} model trained by marking a single word by surrounding it with special tokens as input and then generates the normalized variant as output. We follow the original implementation, but skip their synthetic data augmentation step, as it is non-trivial for our target languages.

\begin{figure}[t]
    \centering
\includegraphics[width=0.45\linewidth]{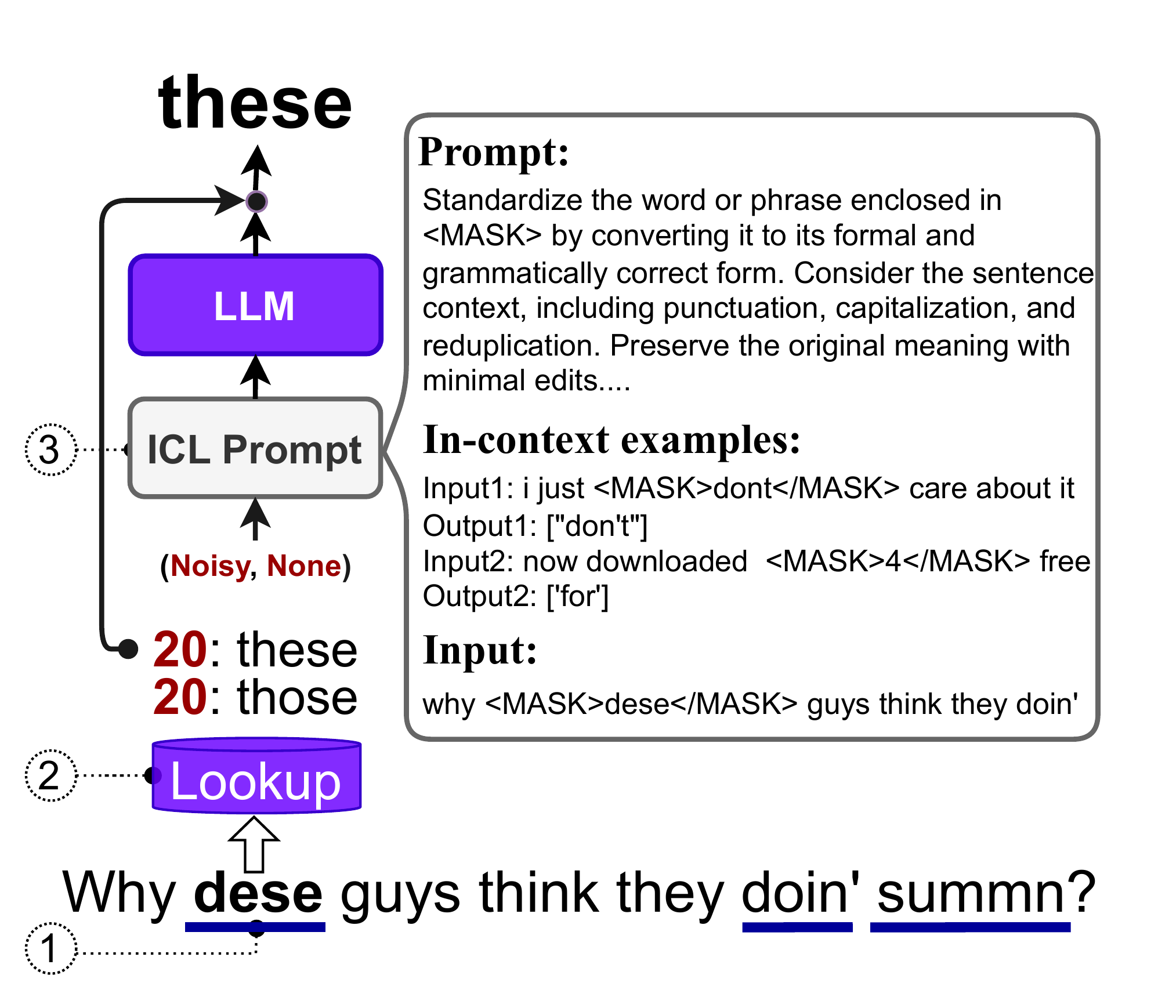}
    \caption{Two-shot In-Context Learning: \protect\circnum{1} \textit{Detection}: An encoder-based model detects typos in informative words. \protect\circnum{2} \textit{Dictionary}: dictionary lookup with Miller-Madow entropy selection \protect\circnum{3} \textit{LLM}: in-context prompt prediction with an LLM.}
    \label{fig:fullpipeline}
\end{figure}

\subsection{LLM-based Pipeline} 

LLMs have shown remarkable performance across a wide variety of NLP tasks in few-shot settings. 
However, our initial empirical results showed that lexical normalization is challenging for these models (results in Appendix~\ref{llmonly}), so we propose a pipeline approach (Figure~\ref{fig:fullpipeline}).
We first use an encoder-based detection model, which is trained to predict which words are in need of normalization (results in Appendix~\ref{appendix:detection_model}).  This is a binary sequence labeling task, for which we use XLM-R~\cite{conneau-etal-2020-unsupervised}.
This makes the use of LLM's more efficient (as they only focus on relevant words), and avoids overnormalization.
After the detection, we use the most frequent replacements from the training data, where we use Miller-Madow entropy for cases with multiple options, and only replace to low-entropy candidates. 
After this, we treat text normalization as an in-context prompting task. We construct an in-context prompt using the detected word and its surrounding context as input. We mark the word to guide the LLM which word to normalize. We also use randomly sampled few-shot instances from the training data to exemplify the task, we empirically choose 8-shot as it showed a good tradeoff 
(see Appendix~\ref{appendix:incontext}).
%
We conduct experiments using four recent LLMs: 
1) \textit{gemma-3-27b-it}~\cite{team2025gemma}, 
2) \textit{gpt-4o}~\cite{hurst2024gpt}, 
3) \textit{Llama-3.3-70B-Instruct}~\cite{grattafiori2024llama}, and 
4) \textit{Qwen2.5-72B-Instruct}~\cite{yang2024qwen2}.

%% file: body/results.tex
\section{Results}

\begin{table}
    \centering
    \setlength{\tabcolsep}{1.5pt} 
    \begin{tabular}{|l|
            S[table-format=2.2] S[table-format=2.2] S[table-format=2.2] S[table-format=2.2] S[table-format=2.2] S[table-format=2.2]|
            S[table-format=2.2] S[table-format=2.2] S[table-format=2.2] S[table-format=2.2] S[table-format=2.2] S[table-format=2.2]|
        }
        \hline
        \multirow{2}{*}{\textbf{Lang.}} & \multicolumn{6}{c|}{\textbf{ERR}} & \multicolumn{6}{c|}{\textbf{F1-score}} \\
        \cline{2-13}
        & \textbf{MFR} & \textbf{ÚFAL} & \textbf{LLaMA3} & \textbf{Qwen2.5} & \textbf{Gemma3} & \textbf{GPT-4o} 
        & \textbf{MFR} & \textbf{ÚFAL} & \textbf{LLaMA3} & \textbf{Qwen2.5} & \textbf{Gemma3} & \textbf{GPT-4o} 
        \\
        \hline
        
        IN & 
        58.94           & \underline{67.94}     & 63.24	         & 64.73	             & 66.45                & \textbf{68.35}       
        & 74.17 & \underline{81.42} & 78.34 & 79.37 & 80.60 & \textbf{81.95}
        \\
        
        JA & 
        5.77            & \textbf{39.24}        & 16.27	         & 20.34	             & 14.66                & \underline{22.52}    
        & 35.42 & \textbf{61.09} & 38.50 & 41.59 & 35.83 & \underline{45.22}
        \\
        
        KO & 
        \textbf{6.35}   & \underline{-2.38}     & -14.82         & -6.88                 & -8.73                & -8.73                
        & 15.71 & \textbf{25.43} & 4.42 & 5.95 & \underline{8.17} & 7.74
        \\
        
        TH & 
        41.87           & 6.90                  & 40.39	         & 42.23	             & \underline{43.09}    & \textbf{43.95}       
        & 66.41 & 24.73 & 67.36 & 68.34 & \underline{69.04} & \textbf{69.80}  
        \\
        
        VI & 
        75.58           & 77.35                 & 76.41	         & \underline{77.79}     & 77.57                & \textbf{80.04}       
        & 86.49 & 87.63 & 87.21 & 88.02 & \underline{87.91} & \textbf{89.31}
        \\  \hline
        
        DA & 
        49.68           & 65.19                 & 64.56	         & 61.50	             & \underline{65.82}    & \textbf{66.14}       
        & 67.88 & 80.63 & 80.62 & 78.84 & \underline{81.42} & \textbf{81.81}  
        \\
        
        DE &
        31.95           & \textbf{61.60}        & 48.12	         & 51.13	             & 53.45                & \underline{55.03}    
        & 51.25 & \textbf{78.43} & 67.83 & 70.22 & 71.86 & \underline{73.09} 
        \\
        
        EN & 
        64.93           & \textbf{70.40}        & 68.56	         & \underline{69.78}	 & 69.22                & 69.42                
        & 79.79 & \textbf{84.21} & 82.94 & \underline{83.67} & 83.36 & 83.51
        \\
        
        ES 
        & 25.57           & 42.20                 & 40.33	         & 45.74	             & \underline{47.26}    & \textbf{48.93}       
        & 40.92 & 64.36 & 63.41 & 67.60 & \underline{68.77} & \textbf{70.36}
        \\
        
        HR
        & 36.52           & \textbf{61.35}        & 45.11	         & 45.16	             & 51.83                & \underline{55.12}    
        & 55.32 & \textbf{77.60} & 66.91 & 66.65 & 71.73 & \underline{74.30}
        \\
        
        ID-EN
        & 61.17           & \textbf{65.46}        & 64.66	         & \underline{65.29}     & 64.49                & 64.49                
        & 77.93 & \textbf{82.57} & 81.90 & \underline{82.19} & 81.84 & 81.91
        \\
        
        IT & 
        18.63           & \underline{46.53}        & 39.93	         & 44.55	             & 44.88                & \textbf{46.86}    
        & 40.85 & \textbf{72.45} & 63.87 & 68.06 & 67.57 & \underline{69.91}
        \\
        
        NL &
        37.58           & \textbf{56.96}        & 42.70	         & 44.39	             & 50.54                & \underline{56.37}    
        & 55.89 & \underline{73.71} & 62.58 & 64.12 & 69.30 & \textbf{73.83}
        \\
        
        SL &
        56.71           & \textbf{71.98}        & 60.30	         & 60.51	             & 66.55                & \underline{70.90}    
        & 73.43 & \textbf{84.82} & 77.03 & 77.14 & 81.20 & \underline{84.03}
        \\
        
        SR &
        42.52           & \textbf{64.48}        & 54.07	         & 51.68	             & 59.58                & \underline{62.26}    
        & 60.68 & \textbf{79.69} & 72.45 & 70.55 & 76.38 & \underline{78.27}
        \\
        
        TR &
        15.42           & \textbf{57.60}        & 40.26	         & 38.01	             & 50.84                & \underline{54.90}    
        & 28.33 & \textbf{75.51} & 61.60 & 59.15 & 69.68 & \underline{73.45}
        \\
        
        TR-DE &
        21.82           & \textbf{58.19}        & 35.64	         & 37.45	             & 42.10                & \underline{53.37}    
        & 40.85 & \textbf{76.19} & 58.66 & 60.43 & 64.49 & \underline{73.14}
        \\
        \hline
        
        \textbf{Avg.} & 
        38.29           & \textbf{53.59}        & 46.22	         & 47.85	             & 50.57                & \underline{53.53}    
        & 55.96 & \underline{71.20} & 65.62 & 66.58 & 68.77 & \textbf{71.27}   
        \\
        \hline
    \end{tabular}
    \caption{ERR scores and F1 scores across language settings. Bold indicates the highest (best) score per row for each metric; underlined denotes the second highest. LLM results are averaged over three runs each.}
    \label{tab:mainresults}
\end{table}

Following the original MultiLexNorm~\cite{van-der-goot-etal-2021-multilexnorm}, we use Error Reduction Rate (ERR) as our main evaluation metric. ERR represents the \% of the task that is solved, it normally ranges from 0-100 (higher is better), and a negative ERR indicates more incorrect normalizations than correct normalizations.
In addition to ERR, we report F1 scores. As van der Goot observes (Section 5.1, p. 76) \cite{van2019normalization}, F1 in lexical normalization is ambiguous, since tokens normalized in the gold standard but mapped to the wrong word may be counted as false positives or false negatives. In this paper, we treat them as false positives.
Results (Table~\ref{tab:mainresults}) show that languages that use the Latin script exhibit higher ERR and F1 scores than those using non-Latin scripts. 
LLMs show promising performance considering that they have not been fine-tuned for this task, but the open models still underperform compared to UFAL for most languages. However, GPT-4o matches UFAL in overall performance and outperforms it on Thai, Vietnamese, and Indonesian. 
UFAL underperforms especially on Thai and Korean, presumably because its byte-level representations are inefficient for these scripts (Section~\ref{ap:tokenization}), confirming previous work reporting low performance for these languages on other NLP tasks~\cite{edman-etal-2024-character, xue-etal-2022-byt5}.
Performance on Japanese and Korean is low overall, we provide an analysis of the remaining challenges in the following section.

%% file: body/analysis.tex
\section{Analysis} 
%
To gain a better understanding of current state-of-the-art bottlenecks, we provide an automatic and manual analysis of different aspects of model outputs.
\subsection{Finding the right normalization}

\begin{figure}
    \centering
    \begin{subfigure}[b]{0.33\textwidth}
        \centering
        \includegraphics[width=\textwidth]{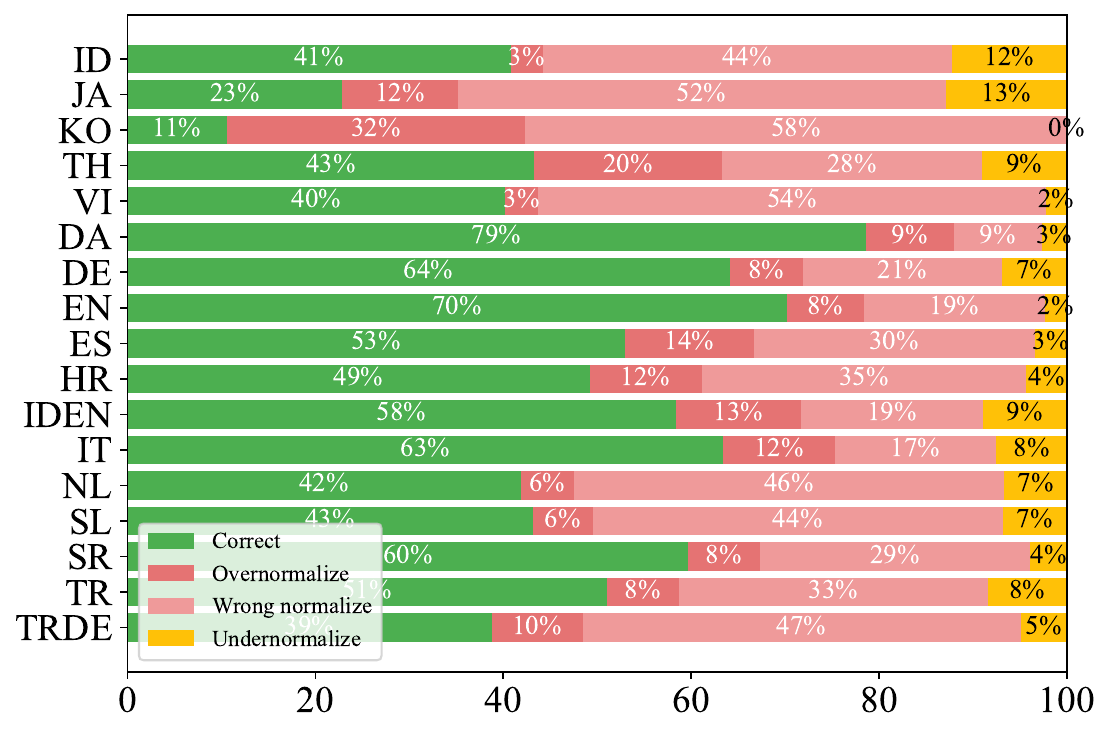}
        \caption{Llama3.3}
    \end{subfigure}
    \hspace{0.01\textwidth}
    \begin{subfigure}[b]{0.33\textwidth}
        \centering
        \includegraphics[width=\textwidth]{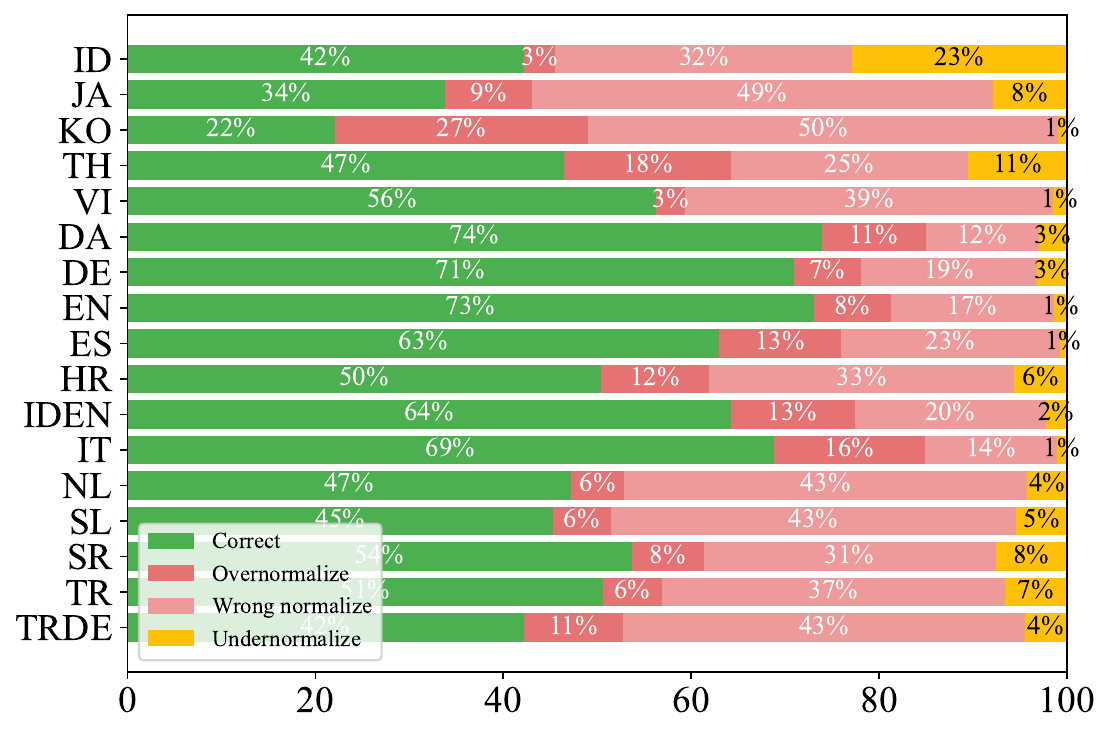}
        \caption{Qwen2.5}
    \end{subfigure}
    \begin{subfigure}[b]{0.33\textwidth}
        \centering
        \includegraphics[width=\textwidth]{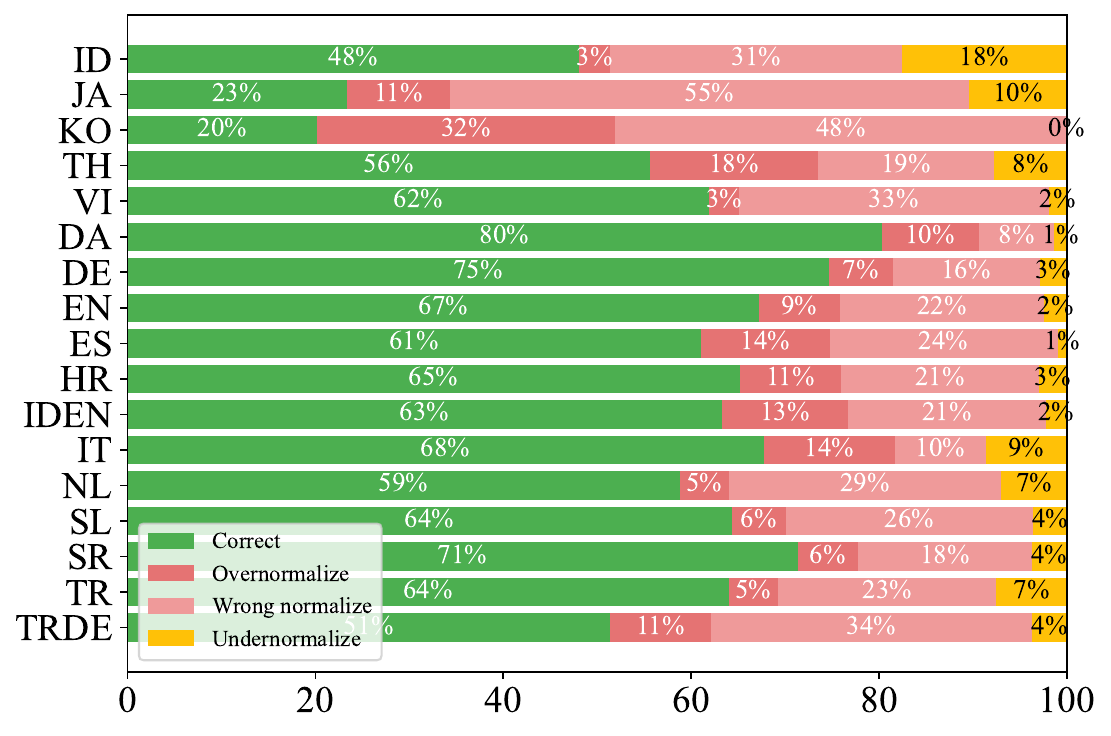}
        \caption{Gemma3}
    \end{subfigure}
    \hspace{0.01\textwidth}
    \begin{subfigure}[b]{0.33\textwidth}
        \centering
        \includegraphics[width=\textwidth]{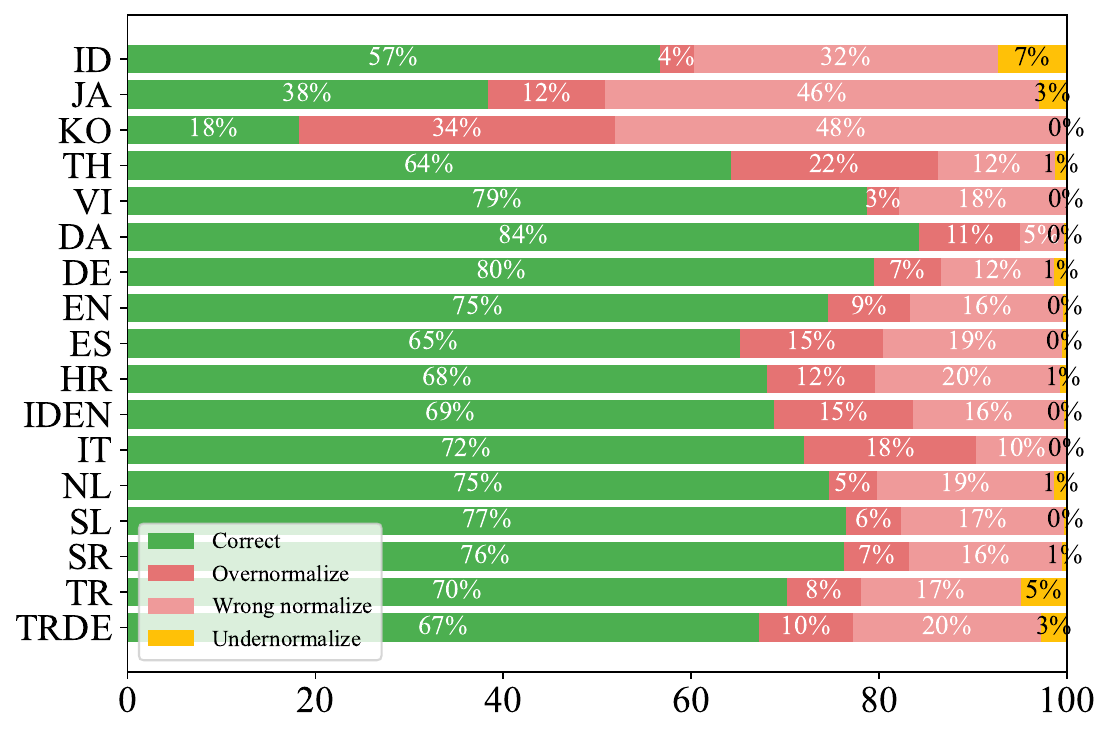}
        \caption{GPT-4o}
    \end{subfigure}

    \caption{Normalized prediction distributions across languages for each LLM.}
    \label{fig:llm_comparison}
\end{figure}

\label{app:categories}
We categorized the LLM predictions into four types according to which step the model fails. 
It is clear that normalizing to the wrong normalization candidate is the largest error category (Figure~\ref{fig:llm_comparison}), followed by overnormalization (i.e., replacing a word that is not annotated for replacement) and undernormalization (word needs normalization, models do not normalize).
Overnormalization is the only category that does not increase when comparing the new languages versus the original languages, this is most likely due to the separate detection step.

As shown in Figure~\ref{fig:llm_comparison}, we compared LLM models across different languages.
The results indicate that GPT-4o shows the lowest error distribution. Followed by Gemma3, where the under-normalized error category increases for Gemma3 compared to GPT-4o. LLama3 and Qwen3 show similar error rates across languages.
From our inspection of the under-normalized error, we found that the models tend to refuse to generate the answer or not follow the instruction. We hypothesize that informal words confuse the models.

\subsection{Human error categorization}
\begin{figure}[H]
    \centering
    \includegraphics[width=0.33\linewidth]{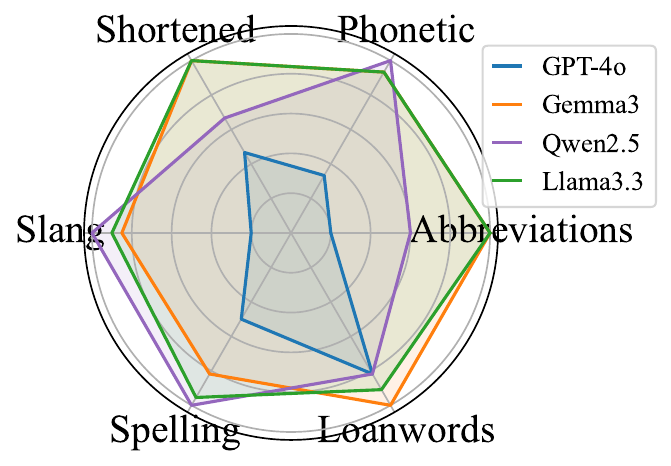}
    \caption{Error rates normalized by the maximum count in each LLM error category.}
    \label{fig:err_type_dis}
\end{figure}
We sampled 178 instances, which were detected as words that need normalization (covering Vietnamese, Thai, and Indonesian; we left out Korean and Japanese due to their low performance) and categorized them into six normalization sub-categories. 
Performance on these sub-categories (Figure~\ref{fig:err_type_dis}) shows that language models have distinctive patterns in their errors (except Llama3.3 and Gemma3, which are quite similar).
The main remaining errors for GPT-4o are spelling errors and transliterated loanwords. Qwen2.5 is relatively good at shortened words and abbreviations, whereas the other models have a relatively uniform distribution. We hypothesize that this is an effect of their pre-training data, which might include different varieties of phenomena.

\subsection{Analysis of segmentations}
\label{ap:tokenization}
To gain a deeper understanding of the low performance of the new languages in the UFAL model, which is based on the ByT5 language model, we examine the effect of the input units on final performance by inspecting their sizes relative to word boundaries and UTF-8 characters in our data.

We report the average number of characters per subword, and the average number of subwords per word for each language. Note that subwords in ByT5 are bytes, and the number of characters per subword is thus negative. We use Gemma-3-27b-it as an example model here, but the trends are highly similar for the other LLM's. 

The statistics (Table~\ref{tab:char_subw_word}) show very clear trends. For language in non-Latin scripts, the number of characters per subword is extremely low for ByT5, and also substantially lower for Gemma-3-27b-it compared to Latin-script languages. A similar effect is seen on the number of subwords per word, except that mainly Korean and Thai are outliers for ByT5, and Korean and Indonesian are outliers for Gemma-3-27b-it, which is probably an effect of the settings of the training of their tokenizer.

\begin{table}
    \centering
    \begin{tabular}{l 
                    S[table-format=2.2] S[table-format=2.2] 
                    S[table-format=2.2] S[table-format=2.2]
                     c} 
        \toprule
        \multirow{2}{*}{\textbf{Languages}} 
        & \multicolumn{2}{c}{\textbf{byt5-base}} 
        & \multicolumn{2}{c}{\textbf{gemma-3-27b-it}} 
        & \multirow{2}{*}{\textbf{Script}} 
        \\
        \cmidrule(lr){2-3} \cmidrule(lr){4-5}
        & \textbf{char/subw} 
        & \textbf{subw/word} 
        & \textbf{char/subw} 
        & \textbf{subw/word} \\  
        \midrule
        IN     & 1               & 5.43            & 2.48           & 2.19   & Latn \\
        JA     & 0.34            & 5.11            & 1.42           & 1.23   & Jpan \\
        KO     & 0.34            & 9.61            & 1.13           & 2.87   & Kore \\
        TH     & 0.36            & 9.51            & 2.36           & 1.47   & Thai \\
        VI     & 0.72            & 4.24            & 1.73           & 1.77   & Latn \\ \midrule 
        DA     & 0.97            & 4.34            & 2.39           & 1.77   & Latn  \\
        DE     & 0.99            & 4.56            & 2.76           & 1.63   & Latn  \\
        EN     & 1               & 4.62            & 2.59           & 1.78   & Latn  \\
        ES     & 0.99            & 4.85            & 2.48           & 1.93   & Latn  \\
        HR     & 0.98            & 4.56            & 2.36           & 1.89   & Latn  \\
        ID-EN  & 1               & 4.61            & 2.65           & 1.74   & Latn  \\
        IT     & 0.99            & 4.83            & 2.52           & 1.9    & Latn  \\
        NL     & 1               & 4.35            & 2.42           & 1.79   & Latn  \\
        SL     & 0.98            & 4.08            & 2.28           & 1.75   & Latn  \\
        SR     & 0.98            & 4.19            & 2.34           & 1.75   & Latn  \\
        TR     & 0.94            & 6.49            & 2.34           & 2.62   & Latn  \\
        TR-DE  & 0.95            & 5.67            & 2.63           & 2.05   & Latn  \\
        \bottomrule
    \end{tabular}
    \caption{Character per subword and subword per word ratios.
    \label{tab:char_subw_word}}
\end{table}

\subsection{Component Contribution Analysis}

\begin{table}[h!]
\centering
\begin{tabular}{lcccc}
\toprule
Configuration & Llama3.3 & Qwen2.5 & Gemma3 & GPT-4o \\
\midrule
Detection Upper-bound & \multicolumn{4}{c}{85.59} \\
\midrule
LLM Only          & 32.36 & 36.20 & 41.91 & 49.13 \\
+ Lookup          & 46.22 & 47.85 & 50.57 & 53.53 \\
\bottomrule
\end{tabular}
\caption{The average of the ERR scores from all languages of the ablation study that shows the contribution of lookup and LLM, with detection-only as the upper bound.}
\label{tab:component_analysis}
\end{table}

To better understand the contribution of each component in our framework, we conduct an ablation analysis and report its respective performance. 

\paragraph{Span Detection.}  
As shown in Table~\ref{tab:component_analysis}, the span detection module is introduced to reduce unnecessary LLM inference. Since only a small portion of the text requires normalization (on average 15.62\% of tokens), the detection model allows us to avoid sending all tokens to the LLM. Concretely, applying span detection reduced the total token count by 95.25\%, from 77,833,519 tokens (all 8-shot input and output prompt tokens) to just 3,693,437 tokens. This substantially reduces the computational burden and makes the framework more practical.  

\paragraph{Lookup Table.}  
We also introduce a lookup table constructed from training data to further improve performance. This component complements the LLMs by directly resolving common cases, resulting in consistent gains across all evaluated models.  

\paragraph{Overall Results.}  
Table~\ref{tab:component_analysis} summarizes the results. The detection model alone provides an effective normalization signal, yielding an upper bound of 85.59 average ERR score. In contrast, LLMs used without additional components perform considerably lower (32.36–49.13). Adding the lookup table leads to substantial improvements across all LLMs.

\subsection{Cost Analysis}
In addition to performance, we analyze the computational cost of our framework.  
Based on GPT-4o API pricing (September 2025: \$2.50 per 1M input tokens and \$10.00 per 1M output tokens), the reduced token count after applying span detection results in a total cost of approximately \$9.3 to run our full multilingual setup. This demonstrates that our framework is both effective and cost-efficient.

%% file: body/data_statement.tex
\section{Data Statement}
\label{app:datastatement}

We provide a dataset statement following the setup proposed by~\citet{bender-friedman-2018-data}:

\begin{enumerate}[label={\Alph*.}]
    \item \textsc{CURATION RATIONALE} \hspace{.2em}
    
\begin{itemize}
    \item Thai: The dataset was collected data from Twitter's platform between 2017 and 2019, where only top trend topics in each month and longest were selected. 
    \item Vietnamese: The corpus was collected from Facebook and TikTok, focusing on diverse content categories and highly engaging public comments.
    \item Indonesian: Extracted from Instagram pages of 104 Indonesian public figures in 2016. Manually filtered for relevance. 
    \item Japanese: The dataset was collected from Twitter's platform between 2010 and 2020.
    \item Korean: The dataset was collected from well known South Korean social media platform ``DC Inside''. 
\end{itemize}

    \item \textsc{LANGUAGE VARIETY} \hspace{.2em}
    We use iso639-3 codes:
\begin{itemize}
    \item Thai: tha
    \item Vietnamese: vie
    \item Indonesian: ind
    \item Japanese: jpn
    \item Korean: kor
\end{itemize}

    \item \textsc{SPEAKER DEMOGRAPHIC} \hspace{.2em} 
\begin{itemize}
    \item Thai: The data was collected based on top-trend Twitter topics. No direct information about the gender distribution or socioeconomic status of the speakers.
    \item Vietnamese: The data was collected from public comments on Facebook and TikTok from a wide range of content categories. No direct information about the gender distribution or socioeconomic status of the speakers.
    \item Indonesian: Collected from Instagram pages. No direct information about the gender distribution or socioeconomic status of the speakers.
    \item Japanese: The data was collected from Twitter posts. No direct information about the gender or socioeconomic status of the speakers.
    \item Korean: The data was collected based on the famous South Korean social media platform ``DC inside'. No direct information about the gender or socioeconomic status of the speakers.
\end{itemize}

    \item \textsc{ANNOTATOR DEMOGRAPHIC} \hspace{.2em} 
\begin{itemize}
    \item Thai: The dataset is annotated by ten linguistic students, and all of them are Thai native speakers aged between 20 to 22 years old.
    \item Vietnamese: The corpus is formulated by six native speakers aged 20–22 with strong Vietnamese language skills and social media experience. Their backgrounds include computer science, Vietnamese studies, economics, and construction.
    \item Indonesian: University students
    \item Japanese: The dataset was annotated by three Japanese university students who are native speakers of Japanese. The first annotator performed sentence normalization, while the second assessed its acceptability and modified it as necessary. Finally, all annotators extracted only lexical normalization, i.e., examples where at least one of the expressions before and after normalization is at the word level.
    \item Korean: The dataset is formulated by two Korean university students, both native Korean speakers.

\end{itemize}

    \item \textsc{SPEECH SITUATION} \hspace{.2em}
\begin{itemize}
    \item Thai: The dataset was created in 2019, where all the texts are from 2017 to 2019, where the topic of each data is discussing influencers, beliefs, goods, government, and daily lives.  
    \item Vietnamese: The dataset was constructed in 2023 using social media comments, reflecting a wide range of public discussions.
    \item Indonesian: The original dataset was created in 2016, but no information about the age of the posts was saved. Normalization annotations were added in 2020.
    \item Japanese: The dataset was created in 2022, where all the texts are from 2010 to 2020, and covers a wide range of topics. Normalization annotations were added in 2024.
    \item Korean: The dataset was created in 2023, the data was extracted from discussions on topics of the platform, including for example: nation, race, politics, social hierarchy, and so on.
\end{itemize}

    \item \textsc{TEXT CHARACTERISTICS} \hspace{.2em} 
    The data is extracted from a variety of social media platforms. Each platform has some platform specific conventions, but also within certain communities there are specific language conventions and uses. For a large portion of the data, a highly informal register is used.

\end{enumerate}

%% file: body/appendix.tex
\section{The effect of in-context samples}
\label{appendix:incontext}
\begin{figure}[H]
    \centering
    \includegraphics[width=0.4\linewidth]{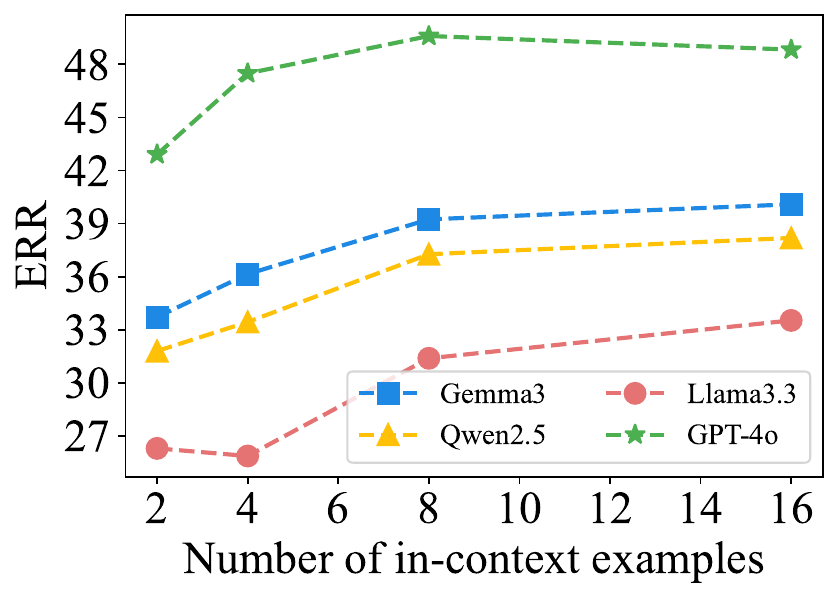}
    \caption{Varying in-context size based on the development set}
    \label{fig:incontext_examples}
\end{figure}
We analyze the effect of in-context examples across five languages (Indonesian, Japanese, Thai, Vietnamese, and English) using four language models.
To evaluate how well the models perform with in-context examples, we vary the number of examples to 2, 4, 8, and 16.
As shown in Fig.~\ref{fig:incontext_examples}, the results show that GPT-4o is the best model, followed by Gemma3, Qwen2.5, and Llama3.3. This trend is consistent across the number of shots.
The performance gains increase for all models when we add more shots. However, performance gains diminish after eight examples.
Therefore, we fix the number of in-context examples to eight for all experiments.